%% file: hml_arxiv.tex
\documentclass{article}

\pdfoutput=1

\usepackage[left=1in, top=1in, bottom=1in, right=1in]{geometry}

\usepackage{amsfonts}
\usepackage{amssymb}
\usepackage{amsmath}
\usepackage{microtype}
\usepackage{graphicx}
\usepackage{subfigure}
\usepackage{booktabs} 
\usepackage{algorithm}
\usepackage{algorithmic}
\usepackage{natbib}
\usepackage{footnote}

\usepackage{amsmath,amssymb,amsthm, braket}
\usepackage{xcolor}

\usepackage{hyperref}

\usepackage[font=small,labelfont=bf]{caption} 

\hypersetup{
    colorlinks=true,
    linkcolor=blue,
    citecolor=blue,
    filecolor=blue,
    urlcolor=magenta,
}

\DeclareMathOperator*{\argmin}{arg\,min}

\DeclareMathOperator{\GS}{\mathcal{GS}}
\DeclareMathOperator{\R}{\mathbb{R}}
\DeclareMathOperator{\LEN}{\mathcal{L}}

\def\E{{\mathbb E}}
\newcommand{\dataset}[1]{{\tt #1}}

\DeclareMathOperator{\indicate}{\mathbf{1}} 

\DeclareMathOperator{\D}{\mathcal{D}}

\DeclareMathOperator{\err}{\textup{err}}

\DeclareMathOperator{\dist}{\textup{dist}}

\def\qed{\vrule height8pt width3pt depth0pt}

\newtheorem{theorem}{Theorem}                               
\newtheorem{lemma}[theorem]{Lemma}                          

\newif\ifarxiv
\newif\ificml
\arxivtrue
\icmlfalse

\allowdisplaybreaks[1]

\title{Metric Learning on Manifolds}
\date{\vspace{-5ex}}
\author{Max Aalto\footnote{Equal contribution} \footnote{Applied Physics and Mathematics Department, Columbia University, New York, US, \texttt{\href{mailto:msa2187@columbia.edu}{msa2187@columbia.edu}}} \and Nakul Verma\addtocounter{footnote}{-1}\footnotemark[\value{footnote}] \addtocounter{footnote}{1}\footnote{Computer Science Department, Columbia University, New York, US, \texttt{\href{mailto:verma@cs.columbia.edu}{verma@cs.columbia.edu}}  } }

\begin{document}

\maketitle

\input{hml_body.tex}

\end{document}

%% file: hml_body.tex
\begin{abstract}
Recent literature has shown that symbolic data, such as text and graphs, is often 
better
represented by points on a curved manifold, rather than in Euclidean space. However, geometrical operations on manifolds are generally more complicated than in Euclidean space, and thus many techniques for processing and analysis taken for granted in Euclidean space are difficult on manifolds. A priori, it is not obvious how we may generalize such methods to manifolds. We consider specifically the problem of distance metric learning, and present a framework that solves it on a large class of manifolds, such that similar data are located in closer proximity with respect to the manifold distance function. In particular, we extend the existing metric learning algorithms, and derive the corresponding sample complexity rates for the case of manifolds. Additionally, we demonstrate an improvement of performance in $k$-means clustering and $k$-nearest neighbor classification on real-world complex networks using our methods.

\end{abstract}

\section{Introduction}
\label{sec:intro}

Training learning models on symbolic datasets, such as text and graphs, generally requires a reasonable representation of said data in an appropriate embedding space.  Moreover, the efficacy of such models is determined in large part by such feature representations -- if the embedding does not accurately reflect the underlying structure of the data, then any analysis performed on the embedding will be correspondingly inaccurate. Predominantly, the embedding space is chosen to be Euclidean and an embedding technique is performed to give the data a corresponding Euclidean vectorial representation. However, recent literature has shown that various data types may have more suitable representations in non-Euclidean spaces. For instance, natural image patches have been shown to be better represented in spaces similar in topology to the Klein bottle \citep{image_patches_on_klein}, 
and network data with hierarchical structure is better represented in hyperbolic spaces \citep{hgeom_poincare_embedding_fb, hgeom_mds_reptradeoff}.


In particular, the recent interest in representations of symbolic data in hyperbolic space has motivated the conversion of several classical machine learning algorithms to operate effectively in such hyperbolic spaces, such as support vector machines \citep{hgeom_svm} and recommender systems \citep{hgeom_recsys}. While these methods may be performed on any set of data embedded in hyperbolic space, they rely crucially on the structure of hyperbolic spaces, such as explicitly given hyperbolic metric tensors, and the corresponding inner products and distance functions. Thus, although the generalization of prediction algorithms to hyperbolic spaces is an important step, a more fundamental problem remains unsolved. Most notably, if a practitioner has identified a new, previously unexplored non-Euclidean representation for their data, they have no way of applying existing algorithms for their data analysis needs. Ideally, there should be a general framework for developing learning algorithms on a large class of non-Euclidean representations. A key challenge, of course, is that this class needs to be broad enough to cover many representations of contemporary interest, while being manageable enough to be conducive to theoretical analysis and efficient algorithm development.

As a first step towards this goal, we consider the problem of distance metric learning. Distance metric learning is a supervised method that has been shown to improve performance of both classification \citep{mlrn_lmnn, mlrn_itml} and clustering \citep{mlrn_mmc} in Euclidean space. However, current methods for distance metric learning are reliant on the convenient algebraic and geometrical structure of Euclidean space, and are not immediately extendable to the more general problem of working on an arbitrary manifold. In this paper, we propose a general framework for distance metric learning on any manifold that is globally diffeomorphic to an open subset of Euclidean space (i.e.\ there exists an atlas for the manifold containing only one chart), which we will henceforth refer to as \emph{generalized surfaces} (or $\GS$). Successful theoretical analysis and practical implementation of distance metric learning algorithms on such manifolds perhaps indicates that this general class is amenable to the design of a wide range of algorithms.

We demonstrate an increase in the quality of clustering and classification on metric learned generalized surfaces. 
Moreover, we derive the corresponding metric learning sample complexity rates for data in such representations, thus directly extending some of the key theoretical results by \citet{mlrn_verma_samplecomplexity} to manifold data.


Additionally we present (i) an algorithm for $k$-means clustering on data embedded on such surfaces, and (ii) an algorithm for approximating the shortest (geodesic) distance on such surfaces. These algorithms were developed during the process of testing the generalized distance metric learning framework, as we require a $k$-means technique for measuring clustering performance on such surfaces and a way to estimate the distances between points on the surface before and after learning the metric. 

\ifarxiv
\begin{figure}[t]
\vskip 0.2in
\begin{center}
\centerline{\includegraphics[width=3.5in]{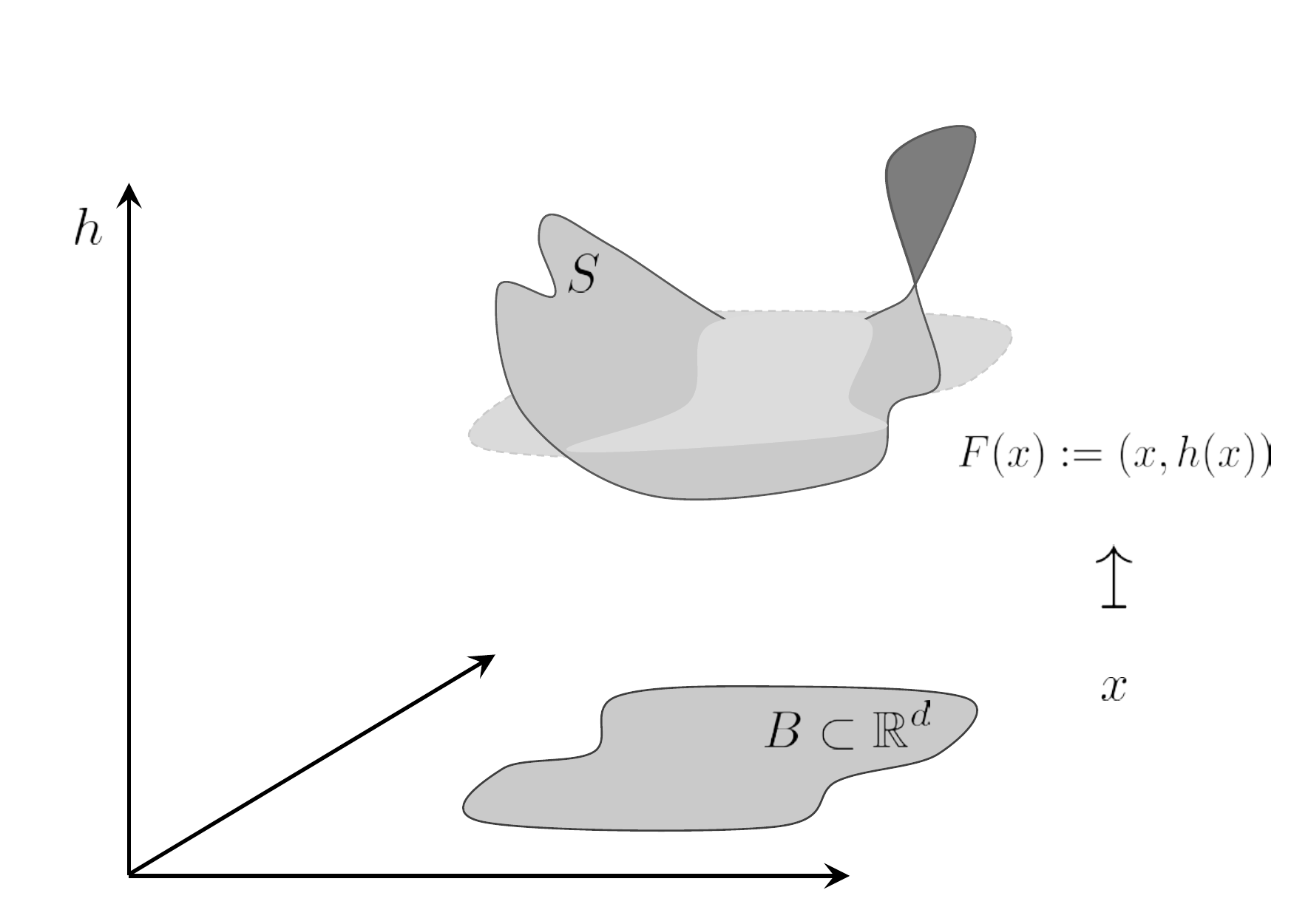}}
\caption{An example $d$-dimensional manifold $S \subset \R^{D}$ that can be expressed in a single chart. The specific map being used here is $F: x \mapsto (x,h(x))$, for some fixed smooth height function $h$. Therefore, the ambient dimension here is $D = d+1$. These types of maps are an important subclass of $\GS$ and are known as a Monge patch parameterization \citep{difgeom_oneill}, or simply, surface manifolds.}
\label{fig:monge_patch}
\end{center}
\vskip -0.2in
\end{figure}
\fi

\section{Formulation}
\label{sec:formulation}
We will focus our attention to a specific class of non-Euclidean representations that can be modelled by manifolds that are globally diffeomorphic\footnote{Two topological spaces are diffeomorphic if there exists a smooth bijection with a smooth inverse between them.} to an open subset of $\mathbb{R}^d$. Equivalently, for any manifold in this class, there exists an atlas that contains exactly one chart. We shall refer to this class of manifolds as generalized surfaces ($\GS$). We'll see later that this restriction of having a single chart representation helps perform explicit geometrical calculations on such surfaces which directly benefits algorithmic design. More concretely, a $d$-dimensional generalized surface $S$ that resides in some ambient space $\mathbb{R}^D$ (not necessarily inheriting the metric structure from the ambient space), is defined by a diffeomorphism $F$ over some $d$-dimensional base space $B\subset \mathbb{R}^d$ that maps it to $\mathbb{R}^D$. Thus, the generalized surface $S \subset \mathbb{R}^D$ is given by the image of the map $F$, i.e.\ $S = { \{F(x): x\in B\} }$. See Figure \ref{fig:monge_patch} for an illustration.

\ificml
\begin{figure}[t]
\vskip 0.2in
\begin{center}
\centerline{\includegraphics[width=3.5in]{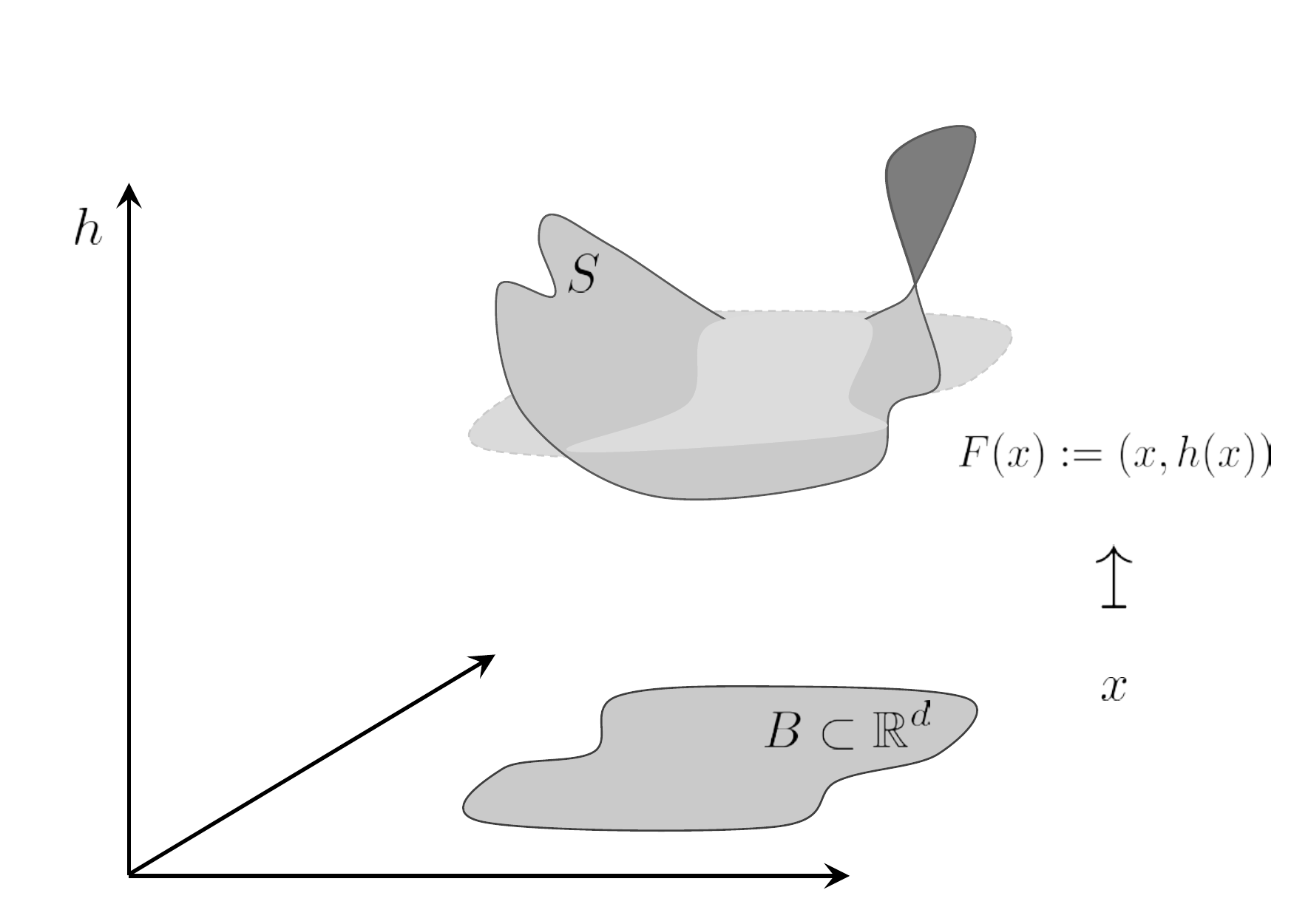}}
\caption{An example $d$-dimensional manifold $S \subset \R^{D}$ that can be expressed in a single chart. The specific map being used here is $F: x \mapsto (x,h(x))$, for some fixed smooth height function $h$. Therefore, the ambient dimension here is $D = d+1$. These types of maps are an important subclass of $\GS$ and are known as a Monge patch parameterization \citep{difgeom_oneill}, or simply, surface manifolds.}
\label{fig:monge_patch}
\end{center}
\vskip -0.2in
\end{figure}
\fi


It is worth noting that this class of manifolds is expressive enough to model non-linear geometries of contemporary interest. For instance, the upper sheet of the hyperboloid of two sheets---an immensely useful model for working in hyperbolic geometry that has recently generated significant interest in machine learning---is one such manifold. Other classical geometries that can be modeled by such a parameterization include elliptical, parabolic, and surface manifolds.

Given such a manifold ${S \in \GS}$, how can we do Maha\-lanobis-type distance metric learning? Naively since ${S\subset \R^D}$, one could potentially consider distance transformations induced by applying linear maps on $\R^D$ itself. Unfortunately, such a map has an undesirable effect 
of distorting the global shape of the representation space $S$ itself. Instead, ideally, what one wants is to have a transformation that can move points around \emph{in} $S$ without having to distort the shape of $S$. 

\begin{figure*}[t]
\vskip 0.2in
\begin{center}
\centerline{
\includegraphics[width=2.3in]{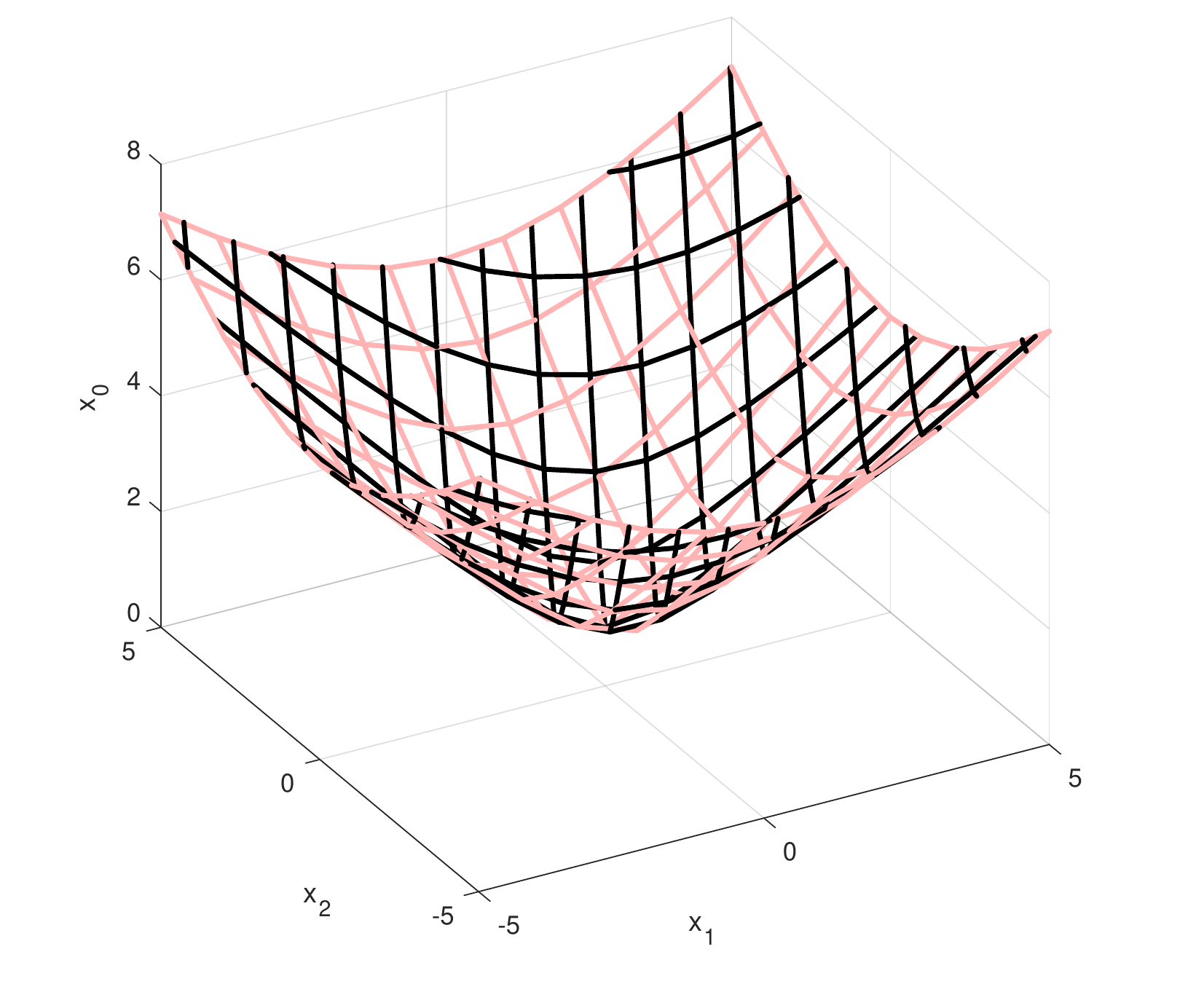} 
\includegraphics[width=2.3in]{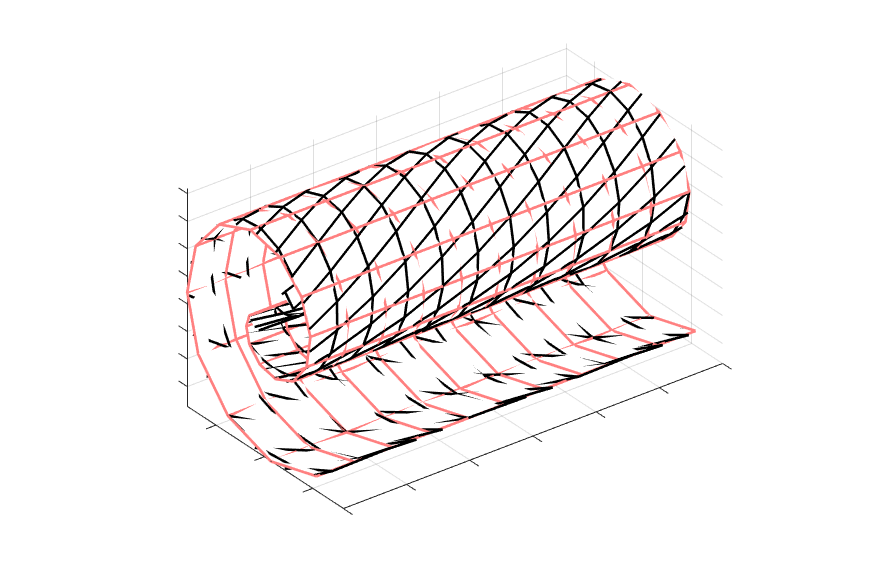} 
\includegraphics[width=2.5in]{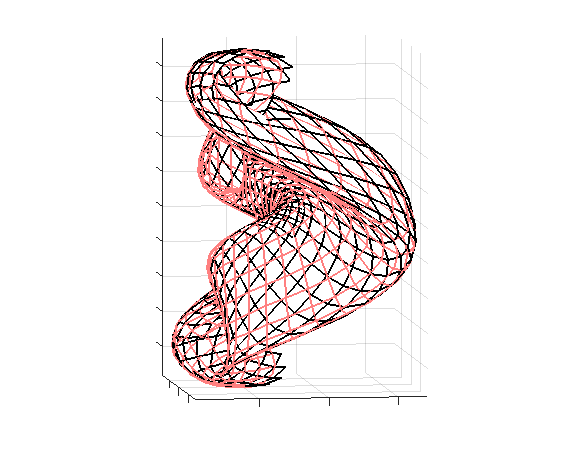} 
}
\caption{Example change in the coordinate system when applying a different $L$ on various manifolds. The red grid lines in lighter shade shows the mapping of the original coordinate system, the black grid lines in the darker shade shows the mapping of the linearly transformed coordinate system. Left: application on a hyperboloid (which uses a non-Riemannian metric structure). Center: application on a `swiss roll' (which has no intrinsic curvature and inherits the Euclidean metric structure from the ambient space). Right: application on a section of a Klein surface (which has intrinsic curvature and inherits the Euclidean metric structure from the ambient space).}
\label{fig:grid_spacing}
\end{center}
\vskip -0.2in
\end{figure*}

This is precisely where our diffeomorphism $F$ comes in handy. Since\footnote{Stating $S=F(B)$ is clearly an abuse of notation; we simply mean $S = \{F(x) \;|\; x\in B\}$. We will make similar abuse of notation throughout the text for sake of clarity and readability.} $S = F(B)$, rather than applying transformations on $S$ directly (which distorts $S$), one can consider applying transformations on the base space $B$ instead. Let $L$ be a linear transform on $\R^d$; then the transformed $S$, namely $S_L$, is defined as $ (F \circ L) (B) = F(L(B)) = F(LB)$. One simple way to understand why this has the desired effect is to imagine a coordinate grid in $B$. Applying $L$ first (linearly) distorts the coordinate grid -- stretching it in some directions and compressing it in others. The subsequent application of $F$ maps this transformed grid into the same shape as $S$. See Figure \ref{fig:grid_spacing} for an illustration. This therefore has the 
requisite effect of stretching some directions in $S$ while compressing other directions in $S$, thus in effect ``pushing" or "pulling" any data that may reside on such a manifold.

While this formulism is helpful, a significant challenge still needs to be addressed. Particularly, since many machine learning algorithms rely on comparing pairwise distances between datapoints, one must still be able to compute distances between points that reside on $S$. Since we allow $S$ to have its own metric structure (and not necessarily using the Euclidean structure that gets inherited from the ambient space), this issue needs to be addressed carefully. We shall provide an algorithmic approximation to calculate distances over the transformed manifolds in Section \ref{sec:dist_over_S}.

%

\subsection{Example Instantiations}
\label{sec:example_instances}

\textbf{Flat manifold.} Taking the base space $B=\mathbb{R}^d$ and choosing $F$ to be the identity map yields a flat surface (i.e.\ Euclidean space). In this case, we recover back the standard Mahalanobis metric learning framework on $d$-dimensional Euclidean spaces. Concretely, let $F(x) = x$ be the identity map. Then, every point $s \in S = F(B) = B$ gets transformed as $F(Lb) = Lb$, where $b = F^{-1}(s)$, which equals $s$. The (squared) distance between two points $s_1, s_2 \in S$ after this transformation simply becomes ${(s_1-s_2)^\mathsf{T} L^\mathsf{T}L (s_1-s_2)}$. The matrix $L^\mathsf{T}L$ is precisely the (quadratic form of) the Mahalanobis metric \citep{mlrn_verma_samplecomplexity}. 

\begin{figure*}[t]
\vskip 0.2in
\begin{center}
\centerline{
\includegraphics[width=2.3in]{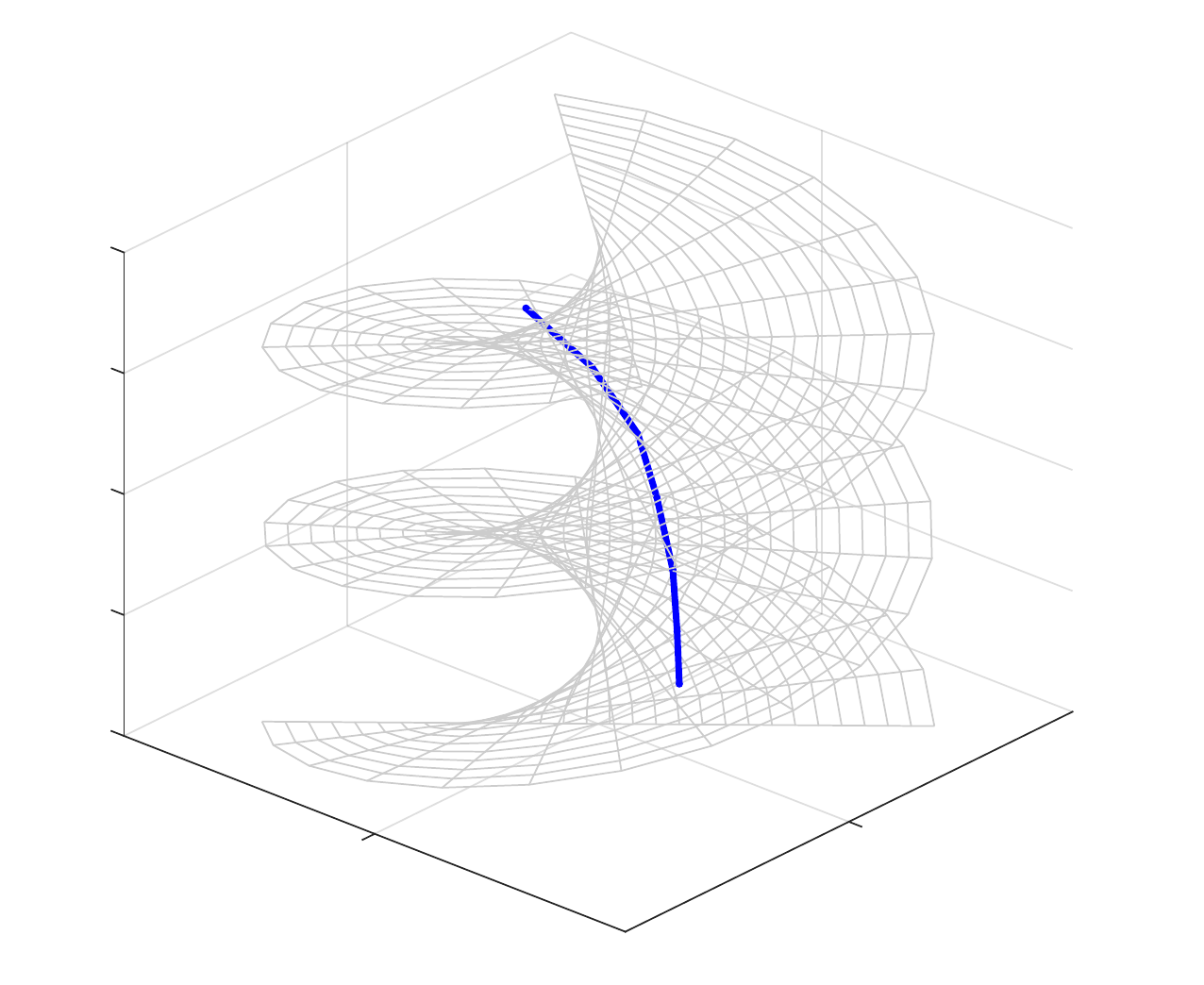} 
\includegraphics[width=2.3in]{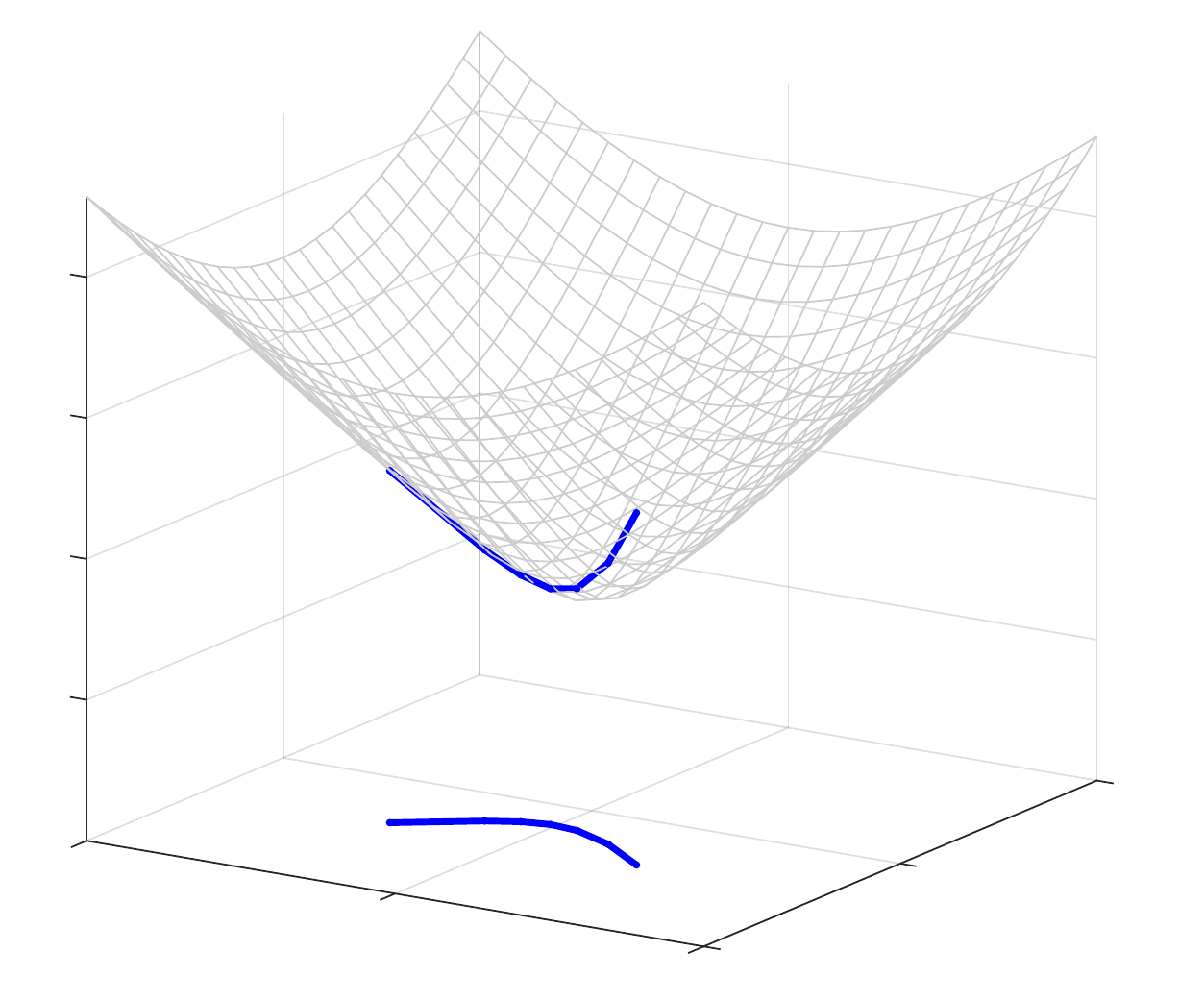} 
\includegraphics[width=2.3in]{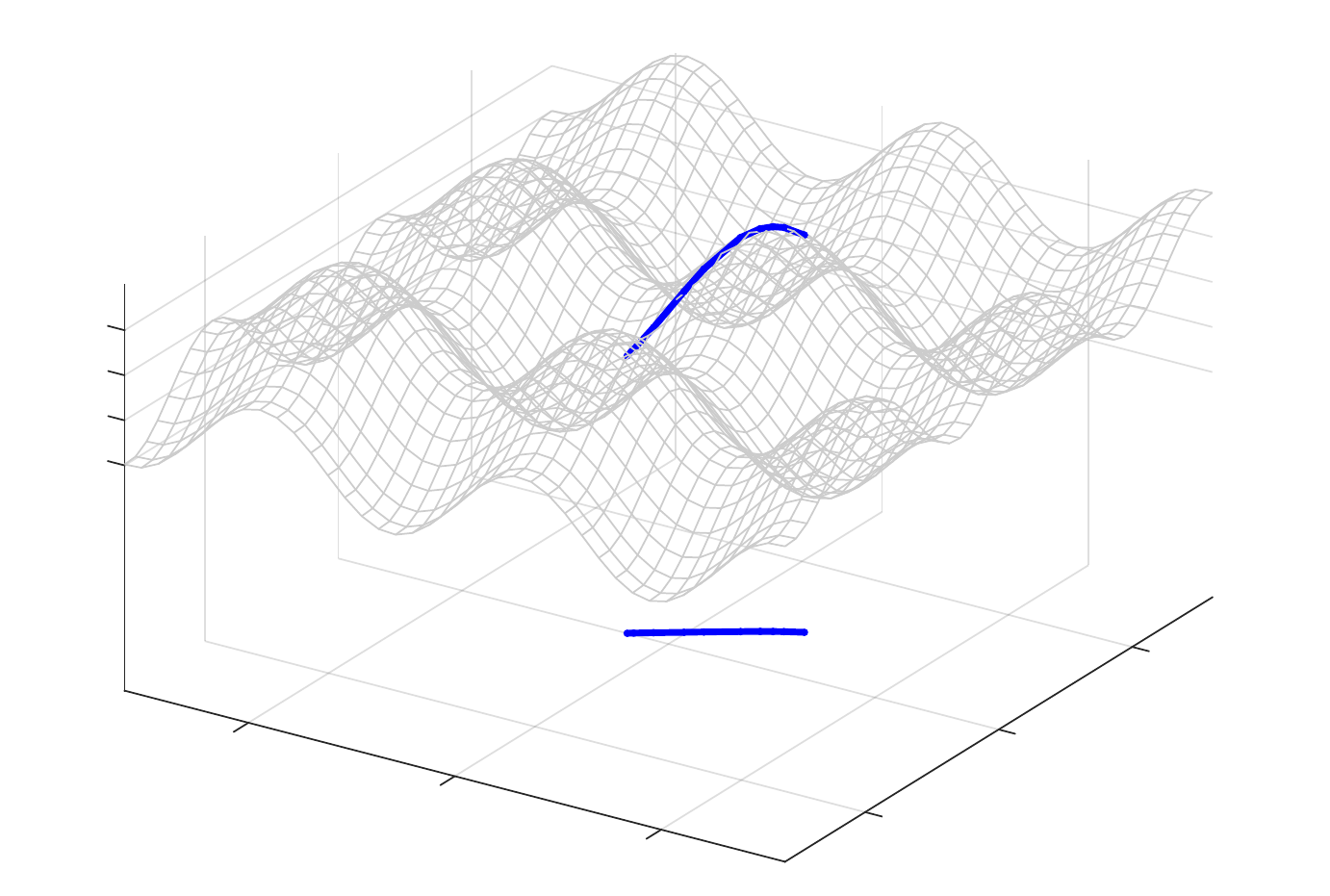} 
}
\caption{Estimation of the shortest path using the proposed algorithm on some example generalized surfaces. Shortest distance approximation on (i) Left: a helicoid -- a generalized surface that inherits ambient Euclidean metric structure, (ii) Center: a hyperboloid -- a surface that does not use the Euclidean (or even a Riemannian) metric structure, (iii) Right: a sinusoid -- a surface that inherits the ambient Euclidean metric structure. Observe that the hyperboloid and sinusoid are surfaces (center and right plots), that is, have a parameterization $x \mapsto(x,h(x))$; we can thus show the corresponding path projection onto the base space (the bottom two coordinates, i.e.\ the bottom plane).}
\label{fig:path_estimation}
\end{center}
\vskip -0.2in
\end{figure*}

\textbf{Hyperboloid manifold.} A classic example of a non-linear surface is the $d$-dimensional hyperboloid, which resides in $d+1$ dimensional ambient space. Using standard derivations (see for example an excellent introduction by \citealp{hgeom_reynolds}) the diffeomporhism $F:\R^d \rightarrow \R^{d+1}$ in this case is\footnote{This setting of $F(x)$ yields a hyperboloid of two sheets: one sheet is obtained from the positive root, and the other sheet from the negative root. For hyperbolic geometry, one usually restricts themselves to one of the sheets; typically the positive sheet.} ${x \mapsto \big(x, (1+x^\mathsf{T}x)^{1/2} \big)}$.
See Figure \ref{fig:path_estimation} (center) for an illustration of the 2-dimensional hyperboloid.

Due to the non-linear geometry, the standard Euclidean distance is no longer the shortest distance between a pair of points (cf.\ Figure \ref{fig:path_estimation} center). Instead, distance $\rho^{\textrm{hyp}}$ between any two points $s_1,s_2$ on the $d$-dimensional hyperboloid $S \subset \mathbb{R}^{d+1}$ is given by 
\begin{align*}
 \rho^{\textrm{hyp}}(s_1,s_2) & := \textup{arccosh}(-s_1^\mathsf{T} G s_2), 
\end{align*}
where the $(d+1) \times (d+1)$ matrix $G :=\Big[\begin{array}{cc} I_{d\times d} & 0\\ 0 & -1 \end{array} \Big]$ encodes the (indefinite) Lorentzian metric signature used for computing the innner product between $s_1$ and $s_2$ (see \citealp{hgeom_reynolds}, for a detailed derivation of distance).

Therefore, the distance between linearly transformed points on the hyperboloid $S$, that is $F(LB)$, simply becomes
(for any $L$)
\ificml
\begin{align*}
\rho^{\textrm{hyp}}_L (s_1,s_2) &=\rho^{\textrm{hyp}} (F(Lb_1),F(Lb_2)) \\
&= \textup{arccosh}\Big( \sqrt{(1+\Delta_{11})(1+\Delta_{22})} -\Delta_{12} \Big), 
\end{align*}
\else
\begin{align*}
\rho^{\textrm{hyp}}_L (s_1,s_2) &=\rho^{\textrm{hyp}} (F(Lb_1),F(Lb_2)) 
= \textup{arccosh}\Big( \sqrt{(1+\Delta_{11})(1+\Delta_{22})} -\Delta_{12} \Big), 
\end{align*}
\fi
where $\Delta_{ij} := b_i^\mathsf{T}L^\mathsf{T}L b_j$, and $b_i = F^{-1}(s_i)$.

\textbf{Helicoid manifold.}
We take the helicoid as our last demonstrative example. This non-linear manifold is \emph{not} a surface (i.e.\ it does not have a mapping of the kind $F: x \mapsto (x,h(x))$, cf.\ Figure \ref{fig:monge_patch}), but does have a global diffeomorphism ${F: \R^2\rightarrow \R^3}$ defined as ${(x_1,x_2) \mapsto \big(x_1 \cos(x_2), x_1 \sin(x_2), x_2\big)}$, and is thus a generalized surface. See Figure \ref{fig:path_estimation} (left) for an illustration. It is interesting to note that this relatively simple and well-known surface, which borrows the Euclidean metric structure from the ambient space, does not have a known closed form expression for distances between pairs of points. This necessitates a procedure to approximate distances between datapoints on a generalized surface $S$ that is specified via $F$. (See next section.)

\subsection{Computing Distances on Arbitrary Generalized Surfaces}
\label{sec:dist_over_S}

Recall that the length of any curve $\gamma:[0,1]\rightarrow \mathbb{R}^{D}$ is given by the arc length integral $\int_0^1 { \sqrt{ \big\langle \frac{d}{dt}\gamma(t), \frac{d}{dt}\gamma(t) \big\rangle} dt} $, where the inner product is with respect to whichever metric structure is endowed on the underlying space. Thus, for a given generalized surface $S$ embedded in $\R^D$ (with its own metric tensor, and not necessarily the one inherited from the surrounding Euclidean space), computing the distance between any two points $x, y \in S$ involves minimizing the functional 
$$\LEN[\gamma(t)] := \int_0^1 \sqrt{ \Big\langle \frac{d}{dt}\gamma(t), \frac{d}{dt}\gamma(t) \Big\rangle} dt$$ 
over all paths $\gamma(t)$ such that $\gamma(t) \in S, \forall t \in [0, 1]$ with $\gamma(0) = x$ and $\gamma(1) = y$ (see for instance \citealp{calcofvar}).

However, since $S$ is diffeomorphic to the base space $B$ (via $F$), any path $\gamma(t)$ such that $\gamma(t) \in S \ \forall t \in [0, 1]$ is given by some path $\kappa(t): [0, 1] \to B$, which is then mapped to $\gamma(t)$ by using $F$. In other words, for any $\gamma(\cdot) \subset S$, $\exists \ \kappa: [0, 1] \to B$ such that $F(\kappa(t)) = \gamma(t)$.

Therefore, formally stated, computing the distance function $\rho^\textup{mfd}(x,y)$ for an arbitrary generalized surface manifold (given by $F$) is equivalent to solving the following variational problem:
$$\inf_{\kappa: [0, 1] \to B} \LEN_F[\kappa] = \inf_\kappa \int_0^1 \sqrt{ \Big\langle \frac{d}{dt}F(\kappa(t)), \frac{d}{dt}F(\kappa(t)) \Big\rangle} dt,$$

such that $ F(\kappa(0)) = x, F(\kappa(1)) = y$ (boundary conditions).

Since the integrand is only a function of $\kappa(t)$ and $\dot{\kappa}(t)$ (where $\dot{\kappa}$ denotes the derivative of $\kappa$ with respect to $t$), finding the minima of $\LEN$ is equivalent to finding the solution $\kappa^\ast$ to the following differential equation, also known as the Euler-Lagrange equation \citep{calcofvar}:
\ificml
\begin{align*}
\frac{\partial}{\partial \kappa} \Bigg [ & \sqrt{ \Big\langle \frac{d}{dt}F(\kappa(t)), \frac{d}{dt}F(\kappa(t)) \Big\rangle} \Bigg] \\
&
= \frac{d}{dt} \frac{\partial}{\partial \dot{\kappa}} \Bigg[\sqrt{ \Big\langle \frac{d}{dt}F(\kappa(t)), \frac{d}{dt}F(\kappa(t)) \Big\rangle} \Bigg], 
\end{align*}
\else
\begin{align*}
\frac{\partial}{\partial \kappa} \Bigg [ & \sqrt{ \Big\langle \frac{d}{dt}F(\kappa(t)), \frac{d}{dt}F(\kappa(t)) \Big\rangle} \Bigg] 
= \frac{d}{dt} \frac{\partial}{\partial \dot{\kappa}} \Bigg[\sqrt{ \Big\langle \frac{d}{dt}F(\kappa(t)), \frac{d}{dt}F(\kappa(t)) \Big\rangle} \Bigg], 
\end{align*}
\fi
with the same boundary conditions.
For an arbitrary generalized surface $S$, computing the optimal minimum-distance geodesic path $\kappa^\ast$ can be computationally prohibitive, and therefore we present an algorithm to approximate this path using piecewise linear paths. Let $F^{-1}(x)=a_0$, $a_1,\ldots, a_n, a_{n+1} = F^{-1}(y)$ be $n$ intermediate points on a path $\kappa$ (where $a_0$ and $a_{n+1}$ are the end points). Define 
$$\sigma(a_i,a_{i+1}) := \int_0^1 \sqrt{ \Big\langle \frac{d}{dt}F(\bar\kappa(t)), \frac{d}{dt}F(\bar\kappa(t)) \Big\rangle} dt,$$
where in this case, $\bar\kappa(t) = (1-t) a_i + t a_{i+1}$, a straight line path $a_i$ and $a_{i+1}$ in the base space $B$.\\ 

\begin{algorithm}[htb!]
\caption{Manifold Distance Approximation}
\label{alg:path_approx}
\begin{algorithmic}[1]
\INPUT $x,y \in S$ (path connected), number of intermediate points $n$, number of samples $m$.
\STATE Let $\{a_i\}, i = 1, \ldots,  n$ be a set of $n$ points linearly spaced\footnotemark ~between $a_0 = F^{-1}(x)$ and $a_{n+1} = F^{-1}(y)$ in $B$
\REPEAT
    \FOR{each intermediate point $a_i$ between $a_0$ and $a_{n+1}$}
    \STATE { $r_i = 2 \cdot \max(\|a_i - a_{i-1}\|, \|a_i - a_{i+1}\|)$}
    \STATE {Let $\{b_j\}, j = 1, \ldots, m$ be points sampled from\footnotemark ~$\mathbb{B}(a_i,r_i) \cap B$ }
    \STATE{Set $j^\ast = \argmin_{j} \ \sigma(a_{i-1}, b_j) + \sigma(b_j, a_{i+1})$}
    \IF{$\sigma(a_{i-1}, b_{j^\ast}) + \sigma(b_{j^\ast},a_{i+1})  < \sigma(a_{i-1}, a_i) + \sigma(a_i, a_{i+1})$}
    \STATE{$a_i \leftarrow b_{j^\ast}$}
    \ENDIF
    \ENDFOR
\UNTIL {convergence}
\STATE \textbf{return} ${\rho^F(x,y) = \sum_{i=0}^n \sigma(a_i, a_{i+1})}$ as the approximated length between $x$ and $y$ on $S$.
\end{algorithmic}
\end{algorithm}
\addtocounter{footnote}{-1}\footnotetext{This initialization assumes (for convenience) that the base $B$ is convex and contains the straight line joining the points. If that is not the case, simply initialize $a_i$ in any reasonable way such that each $a_i \in B$.}
\addtocounter{footnote}{1}\footnotetext{$\mathbb{B}(x,r)$ denotes the ball of radius $r$ centered at $x$.}
Note that if any linear transformation $L$ is being applied to points in the base space $B$ (as needed for distance metric learning, cf.\ Section \ref{sec:formulation}), we can simply apply the same equations on $L \kappa(t)$ instead of $\kappa(t)$ in our computations.

A concrete instantiation of our distance approximation procedure (Algorithm ~\ref{alg:path_approx}) for the hyperboloid manifold is given in Appendix \ref{app:derive_hyperbolic_path_approx}. 
A qualitative demonstration of our Algorithm \ref{alg:path_approx} is shown on various types of generalized surfaces in Figure \ref{fig:path_estimation}. It is instructive to note that, at a cursory glace, for the hyperboloid (Figure \ref{fig:path_estimation} center), it may seem like the estimated path on the surface is \emph{not} the shortest path (a sideways bend on the surface seems shorter than the depicted bottom bend). This mismatch is due to limitations of our intuition: recall that hyperboloid inherits the indefinite Lorentzian metric signature. Since we are used to Euclidean metrics, any attempt to visualize shortest distances in other metrics (especially indefinite metrics) is futile. Fortunately, since we do know
the closed form expression for distances on a hyperboloid (cf.\ Section \ref{sec:example_instances}), we can quantitatively evaluate the approximation returned by  Algorithm \ref{alg:path_approx}, and indeed verify that the displayed path is in fact the shortest (see Section \ref{sec:expt_mfd_dist_approx} for details).

\section{Metric Learning on Manifolds}
\label{sec:mmlrn}
With this mathematical machinery in place, we can trivially generalize existing metric learning algorithms such as Large Margin Nearest Neighbor (LMNN, \citealp{mlrn_lmnn}) and Mahalanobis Metric for Clustering (MMC, \citealp{mlrn_mmc}).   

\subsection{MMC on Manifolds for Improved \texorpdfstring{$k$-means} ~~Clustering}
\label{sec:mfd_mmc}
Given labelled data $(x_1,y_1),\ldots, (x_m,y_m)$, the goal for MMC is to find a linear transformation $L$ that brings data from the same category together while pushing away data from different categories \citep{mlrn_mmc}. This pull-push action has the desired effect of making the category-based clusters in the transformed data representation more pronounced, which can thus be easily recovered by a simple clustering algorithm like $k$-means. This is achieved by constructing two sets of pairs---the \emph{similar pairs} set, which we call $P$, and the \emph{dissimilar pairs} set, which we call $Q$---from the given labelled data. Concretely,   (for all $1\leq i,j \leq m$)
\begin{align*}
    P &:= \{(x_i,x_j) \; | \;  y_i = y_j\}, \\
    Q &:= \{(x_i,x_j)\; | \; y_i \neq y_j\}.
\end{align*}
Then, the following optimization finds the desired transformation:
$$
\min_{L \in \R^{d \times d}} \underbrace{\sum_{(x_i,x_j) \in P} \| Lx_i - Lx_j\|^2}_{\textup{pull term}} - \lambda \underbrace{\sum_{(x_i,x_j) \in Q} \| Lx_i - Lx_j\|^2}_{\textup{push term}},
$$
where $\lambda$ is a hyper-parameter controlling the tradeoff between the pull and the push term.

This can be extended to the manifold case, where the given labelled data $(x_1,y_1),\ldots, (x_m,y_m)$ resides on a known $d$-dimensional generalized surface $S$ (specified by the diffeomorphism $F$). Define $b_i$ in the base space $B\subset\R^d$ as the points $b_i = F^{-1}(x_i)$ (for all $1\leq i \leq m$), and let $\rho^F(x_i,x_j)$ denote the distance between points $x_i$ and $x_j$ on $S$ (with respect to whichever metric tensor is endowed on $S$). Then, as before, the $L$-transformed distance on $S$ is (cf.\ Sections \ref{sec:formulation} and \ref{sec:example_instances}):
\begin{equation}
\label{eq:FLB}
\rho^F_L(x_i,x_j) := \rho^F\big(F(Lb_i),F(Lb_j)\big).
\end{equation}
Therefore, the corresponding manifold MMC optimization simply becomes:
$$
\min_{L \in \R^{d \times d}} \underbrace{\sum_{(x_i,x_j) \in P} \rho^F_L(x_i,x_j)}_{\textup{pull term on } S} - \lambda \underbrace{\sum_{(x_i,x_j) \in Q} \rho^F_L(x_i,x_j)}_{\textup{push term on } S},
$$
where $P$, $Q$, and $\lambda$ are defined as before. 

As discussed earlier, if a closed form expression of the distance function over $S$ is known, one can simply plug in that expression for $\rho$; otherwise, they can use Algorithm \ref{alg:path_approx} to approximate it (cf.\ Section \ref{sec:dist_over_S}).

\subsection{LMNN on Manifolds for Improved Nearest Neighbor Classification}
\label{sec:mfd_lmnn}
LMNN \citep{mlrn_lmnn} can be viewed as a ``localized" version of MMC, where instead of pulling and pushing \emph{all} datapoints that belong to the same and different categories (respectively), it pulls and pushes only those in a local 
neighborhood of a given datapoint. This local action directly helps in improving the $k$-nearest neighbor classification quality. Specifically, the classic formulation works on triples (unlike pairs that get used in MMC) of points. Let $(x_1,y_1),\ldots,(x_m,y_m)$ be a given labelled dataset on a $d$-dimensional generalized surface $S$ (specified by the diffeomorphism $F$).
Then, for any $i$, let the relation $j\sim i$ denote that $x_j$ is a true neighbor of $x_i$ (i.e.\ $y_i = y_j$), and the relation $l\not \sim i$
denote that $x_l$ is an imposter neighbor of
$x_i$ (i.e.\ $y_i \neq y_l$). Then, the LMNN optimization on a manifold is given as

\begin{equation*}
    \min_{L\in\R^{d\times d}} \underbrace{\sum_{i,j\sim i} \rho^F_L\big(x_i,x_j \big)}_{\textup{pull term on }S} + \lambda \underbrace{\sum_{i,j\sim i, l\not \sim i} \Big[1+\rho^F_L\big(
    x_i,x_j\big) - \rho_L^F\big(x_i,x_l \big) \Big]_+}_{\textup{push term on } S},
\end{equation*}
where $[\cdot]_+ := \max(\cdot,0)$ denotes the hinge loss.

Observe that selecting $F$ as the identity map immediately gives us back the classical Euclidean formulation of LMNN (cf.\ Section \ref{sec:example_instances}).

We will demonstrate that metric learned manifold representations of symbolic data (rather than naive Euclidean representation) can yield better clustering and classification results. See Section \ref{sec:experiments} for more details.

\section{Sample Complexity of Manifold Metric Learning}
Here, we will derive PAC-style sample complexity bounds for distance metric learning on generalized surface manifolds.
Given a $d$-dimensional generalized surface $S\subset \R^D$ (specified by the diffeomorphism $F$, that is, $S=F(B)$, for an open set $B\subset \R^d$), we want to find a linear
transformation $L^*$ that minimizes some notion of \emph{error} on data drawn from a fixed unknown distribution $\D$ on $S\times\{0,1\}$:
\begin{eqnarray}
\label{eq:errD}
L^* := \argmin_{L \in \mathsterling} \err(L,\D),
\end{eqnarray}
where $\mathsterling$ is a class of linear transformations under consideration.

A practitioner typically defines \emph{error} in a way that makes the optimization prefer those linear transformations that bring data from same class closer together than those from different classes (see, for instance, how error, or the loss function, is defined for LMNN or MMC, cf.\ Section \ref{sec:mmlrn}). More concretely, following the setup discussed in \cite{mlrn_verma_samplecomplexity}, most generally, $\err(\cdot, \cdot)$ can be defined as
\begin{align*}
\err^\lambda(L,\D) := \E_{\substack{(x,y)\sim \D \\ (x',y') \sim \D}} \Big[ \phi^\lambda\Big( \rho^F_L(x,x'), Y \Big) \Big], 
\end{align*}
where $\phi^\lambda(\dist, Y)$ is a generic distance-based loss function that
computes the degree of violation between distance $\rho^F_L(x,x')$ as defined in Eq.\ \eqref{eq:FLB} and the label agreement $Y := \indicate[y = y']$, and
penalizes it by factor $\lambda$. 
This generalized notion of error incorporates many interesting metric learning losses including MMC and LMNN (see \citealp{mlrn_verma_samplecomplexity}, for a detailed discussion and derivation).

We are interested in how well one can approximate Eq.\ \eqref{eq:errD} if only a finite size i.i.d.\ sample $(x_1,y_1),\ldots,(x_m,y_m)$ from $\D$ is available. 
Specifically, let $Z_m$ denote a size $m$ i.i.d.\ sample from $\D$, and $\err(M,Z_m)$ denote the corresponding \emph{empirical} error. 
We can thus define the empirical risk minimizing transform based on $m$ samples as
$L^*_m := \argmin_L \err(L,Z_m)$, 
and compare its generalization performance to that of the theoretically optimal $L^*$, that is, how
\begin{equation}
 \err(L^*_m, \D) - \err(L^*,\D)
\label{eq:erm_conv}
\end{equation}
behaves as the sample size $m$ grows.

Interestingly, we can derive a good convergence rate for the key expression above, and directly extend ${\textrm{Theorem 1}}$ of \citet{mlrn_verma_samplecomplexity} for the case of
$d$-dimensional generalized surface manifolds\footnote{For readability, we only present the statement of the theorem in the main text. An interested reader should refer to Appendix \ref{app:proofs} for a detailed proof.}.

Particularly, (i) let $C_F$ be a measure of how distances are potentially stretched or changed by the diffeomorphism $F$ and the specific choice of metric tensor endowed on $S$; that is, we have
$\rho^F(F(b),F(b')) \leq C_F \|b - b'\|^2$ 
 for all $b,b' \in B$,
(ii) let $C_L$ be a bound on the quadratic form of the linear transformations being considered, that is,
$C_L := \sup_{L \in \mathsterling} \|L^\mathsf{T}L\|_\textup{fro}$, and (iii)
let $C_B$ be the bound on the support of the distribution $\D$ in the base space $B$, that is, $\|b\|^2\leq C_B$, for any $x\sim \mathcal{D}|_S$ such that $F(b) = x$  (with probability 1). Then, we have the following result.

\begin{theorem}
\label{lm:unif_conv_all}
For any generalized $d$-dimensional surface $S$ (with corresponding diffeomorphism $F$), let $\phi^\lambda$ be a distance-based loss function that is $\lambda$-Lipschitz
in the first argument. Then, with probability at least $1-\delta$ over an i.i.d.\ draw of $m$ samples 
 $Z_m$, we have
\ificml
\begin{align*}
 \sup_{L\in \mathsterling}  \Big[ \err^\lambda  (L,\D) - &
\err^\lambda (L,Z_m) \Big]
\\
&\leq O\left( \lambda C_F C_L C_B \sqrt{ \frac{\ln(1/\delta)}{m}}\right).
\end{align*}
\else
\begin{align*}
 \sup_{L\in \mathsterling}  \Big[ \err^\lambda  (L,\D) - 
\err^\lambda (L,Z_m) \Big]
\; \leq \; O\left( \lambda C_F C_L C_B \sqrt{ \frac{\ln(1/\delta)}{m}}\right).
\end{align*}
\fi
\end{theorem}

\begin{figure}[t]
\vskip 0.2in
\begin{center}
\centerline{
\includegraphics[width=3in]{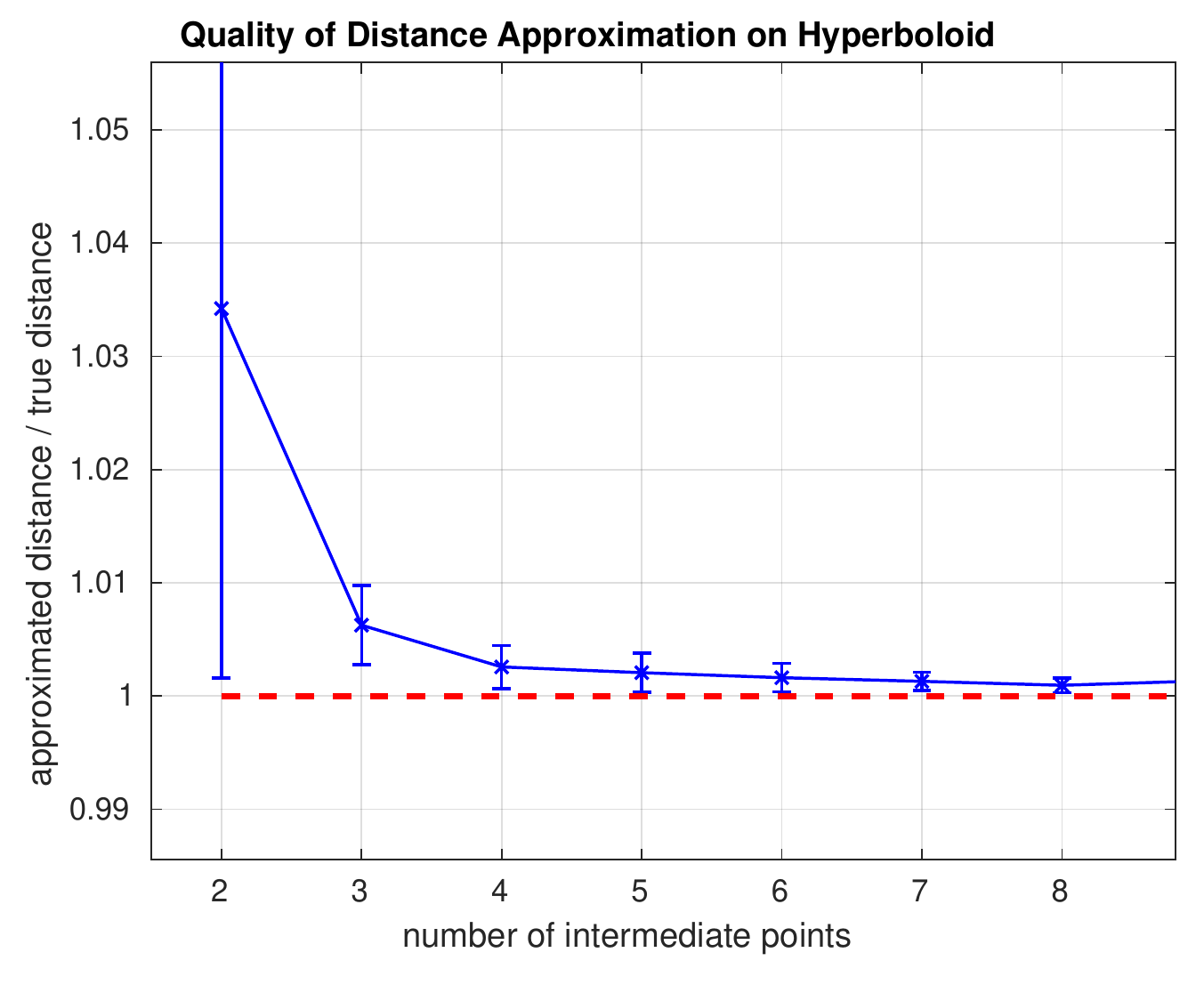}
}
\caption{Quality of distance approximation on a hyperboloid. We plot the ratio between the approximated and the true distance (averaged over multiple pairs of points drawn randomly). The reference line at $1$ (solid dashed line) indicates perfect approximation.}
\label{fig:expt_mfd_dist_approx}
\end{center}
\vskip -0.2in
\end{figure}

The uniform bound presented above this directly implies $1/\sqrt{m}$ rate of convergence of Eq.\ \eqref{eq:erm_conv}. It is instructive to note that while this rate is dimension independent (i.e.\ it is independent of the manifold dimension $d$), the constants (e.g.\ $C_L$) can very well depend on $d$ for some interesting practical cases. 

\begin{figure*}[t]
\vskip 0.2in
\begin{center}
\centerline{
\includegraphics[width=2.0in]{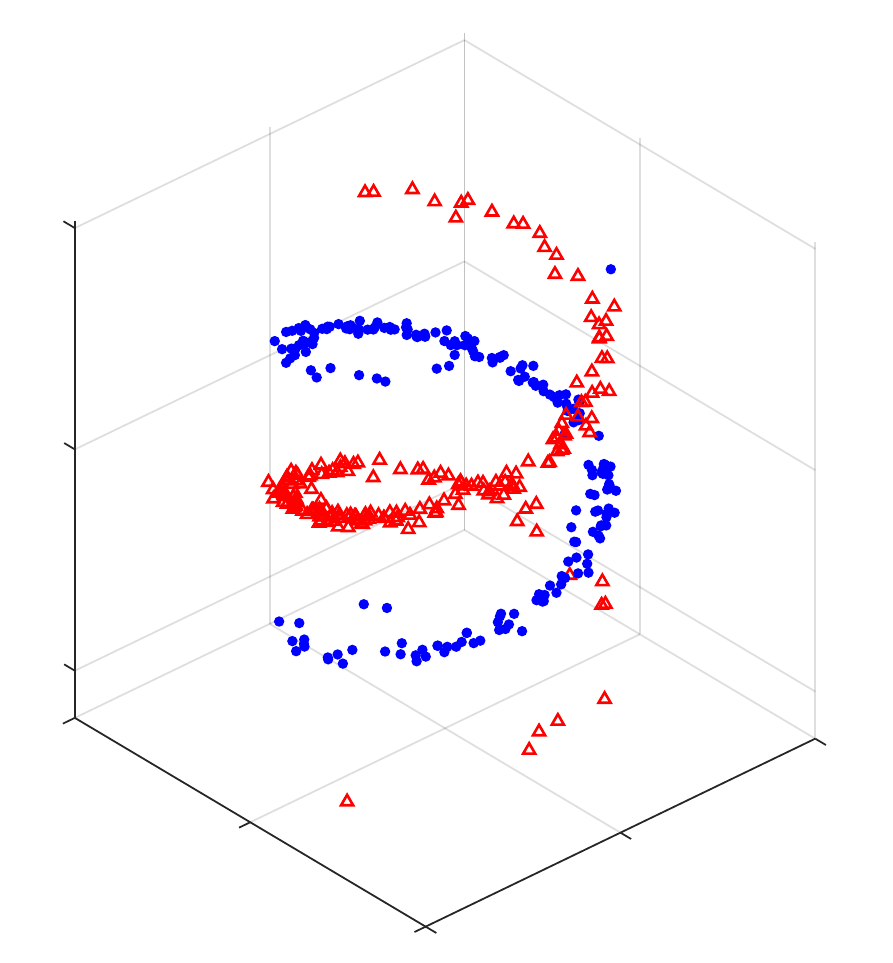} 
\includegraphics[width=2.0in]{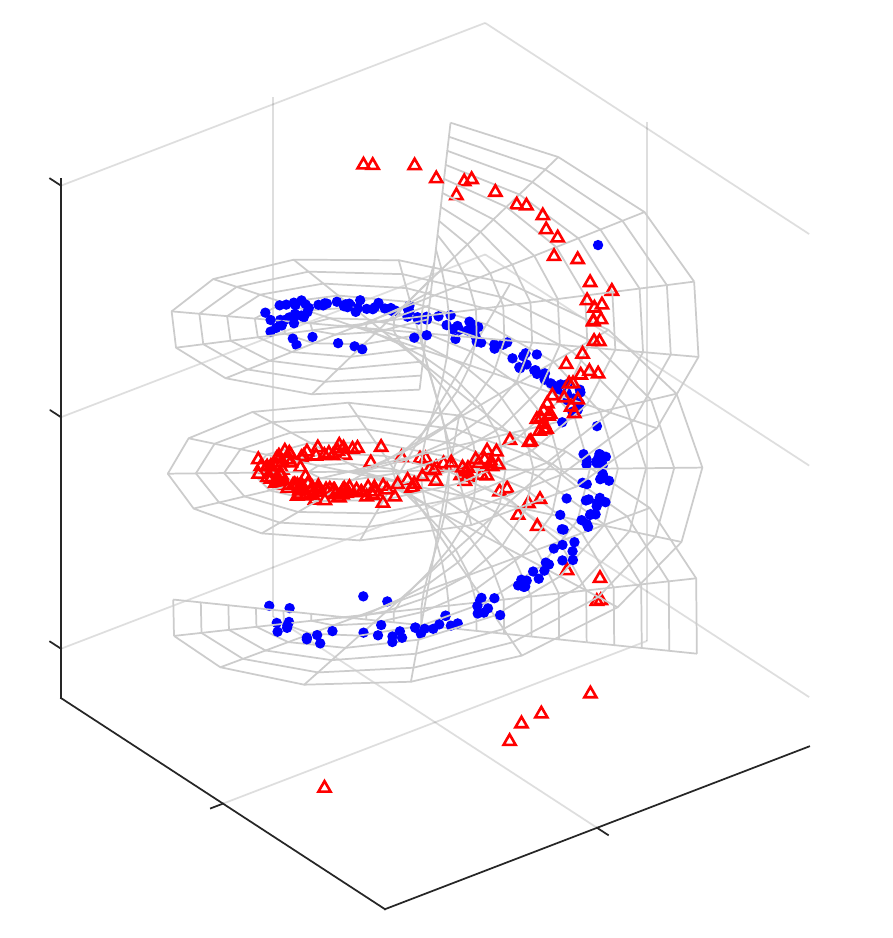} 
\includegraphics[width=2.0in]{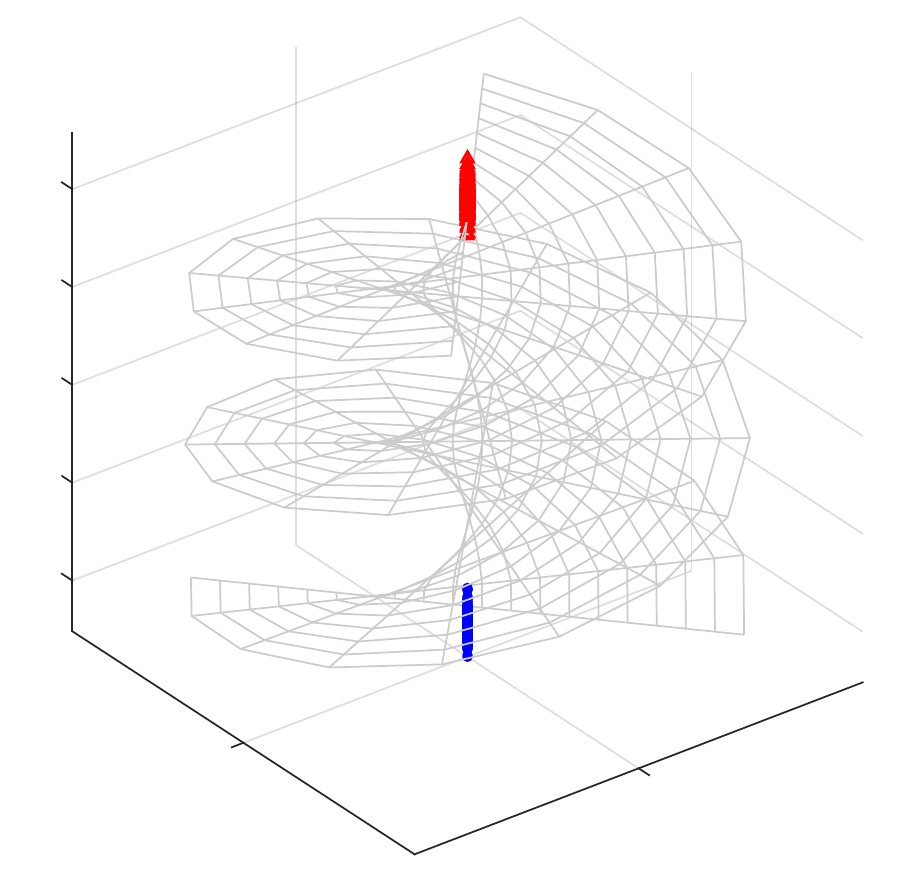}
}
\caption{Clustering result on the helicoid. Left: A synthetic dataset containing two intertwined clusters (cluster identities shown in blue dots and red triangles). Observe that there is no linear transformation that can separate the two clusters for $k$-means. Center: The same synthetic dataset (as depicted in the Left plot) along with the underlying helicoidal structure that can better represent the given dataset. Right: Metric learned representation of the given dataset on the helicoid using manifold-MMC. The two clusters separate out very naturally when an appropriate non-Euclidean representation is considered.}
\label{fig:cluster_helicoid}
\end{center}
\vskip -0.2in
\end{figure*}

\section{Empirical Evaluation}
\label{sec:experiments}

With the manifold metric learning framework in place, we would like to know how much improvement in performance (if any) one can expect by doing metric learning on data that can be modelled more effectively as a generalized surface. Here we compare the performance of both $k$-nearest neighbor classification and $k$-means clustering on representative benchmark datasets. 
Each dataset has a symbolic representation (i.e.\ only the relationships between pairs of datapoints are available), which can be used to embed it (via multidimensional scaling) in any generalized surface (including Euclidean space). We can thus compare prediction performance on Euclidean, metric learned Euclidean, generalized surface, and  metric learned generalized surface representations of the given data\footnote{Code is available at: \texttt{\href{https://github.com/m-k-S/manifold\_ml}{https://github.com/m-k-S/manifold\_ml}}.}.
It is worth noting that even though most of our reported experimental results are on hyperbolic spaces (due to results in previous literature showing it to be an effective representation for such datasets), our published code works for any generalized surface.


\subsection{Approximation of Manifold Distances}
\label{sec:expt_mfd_dist_approx}

As detailed in Section \ref{sec:dist_over_S}, even simple manifolds can have distance functions with no explicit closed form expression. It is imperative to have a good algorithm for approximating distances on generalized surfaces. 
Taking the hyperboloid manifold (which is endowed with the Minkowski metric), which has a known closed form expression for distance, as a reference (cf.\ Section \ref{sec:example_instances}), we can gauge the effectiveness of our proposed Algorithm \ref{alg:path_approx} for distance approximation. 
Figure \ref{fig:expt_mfd_dist_approx} depicts the quality of distance approximation on the hyperboloid as a function of the number of intermediate points used in the computation; as expected, a larger number of intermediate points yields a closer approximation to the true distance. Interestingly, the result also indicates that we can get a good approximation with only a few intermediate points; thus potentially gaining on some computational savings. 


\subsection{Evaluation Setting}

Experiments were conducted on demonstrative synthetic datasets, as well as the following publicly available real-world datasets. \begin{itemize}
    \item \dataset{football}: A network of collegiate American football teams, where edges between nodes represent regular season games in the fall of 2000. There are 12 unique categories, representing the 12 divisions of the NCAA DI football conference \citep{football}.
    \item \dataset{polbooks}: A network of books about US politics published around the 2004 presidential election. Edges between books represent frequent co-purchasing of books by the same buyers on the website Amazon.com. There are 3 unique categories, representing political affiliations (`liberal', `conservative', or `neutral') \citep{polbooks}.
    \item \dataset{adjnoun}: A network of words (nouns and adjectives) taken from the Charles Dickens novel \textit{David Copperfield}. Edges between nodes represent adjacencies between nouns and adjectives, and each node is labelled as either a noun or an adjective \citep{adjnoun}.
    \item \dataset{20newsgroup}: A network of newsgroup documents, where edges between nodes represent a categorical relation. There are 20 unique categories, each representing a different newsgroup \citep{20newsgroup}.
\end{itemize}
\ificml
\begin{table*}[htb]
    \centering
        \caption{$k$-nearest neighbor classification results.}
    \label{tab:clf_res}
    \begin{tabular}{|c|c|c|c|c|}
        \hline 
        Dataset & Euclidean & Euclidean+Metric Learn & Hyperbolic & Hyperbolic+Metric Learn \\
        \hline \hline
 \dataset{football} & 0.41 $\pm$	0.01 & 0.40 $\pm$ 0.09 & 0.29 $\pm$ 0.09 & \textbf{0.25 $\pm$ 0.10} \\
  \dataset{polbooks} & 0.24 $\pm$	0.05 & 0.31 $\pm$ 0.12 & 0.25 $\pm$ 0.06 & \textbf{0.23 $\pm$ 0.06} \\
  \dataset{adjnoun} & 0.58 $\pm$	0.06 & 0.56 $\pm$ 0.07 & 0.55 $\pm$ 0.09 & \textbf{0.49 $\pm$ 0.05} \\
\hline
    \end{tabular}
\end{table*}
\fi
Each of these real-world examples are network-type data, and thus are well-suited to hyperbolic embeddings \citep{hgeom_mds_reptradeoff}. Therefore, we shall use 
hyperboloid as our non-Euclidean representation for 
these network-type datasets.

%

\ifarxiv
\begin{algorithm}[htb!]
\caption{$k$-Means on Generalized Surfaces}
\label{alg:cluster}
\begin{algorithmic}[1]
\STATE $\Theta \leftarrow$ randomly assign each point on the manifold in the dataset $X$ to one of $k$ clusters
\REPEAT
    \FOR{$x_i \in \{x_1, ..., x_n\}$}
    \STATE {current cost = $C(\Theta; X)$}
    \STATE {min cost $\leftarrow$ current cost}
    \FOR{$j \in \{1, ..., k\}$}
    \STATE{Set $\Theta_{new}$ to be equal to $\Theta$ but with the label of $x_i$ set to $j$}
    \STATE{new cost = $C(\Theta_{new}; X)$}
    \IF {new cost $<$ min cost}
    \STATE{$\Theta \leftarrow \Theta_{new}$}
    \ENDIF 
    \ENDFOR
    \ENDFOR
\UNTIL {convergence}
\STATE \textbf{return} $\Theta$
\end{algorithmic}
\end{algorithm}
\fi

\subsection{Clustering on Generalized Surfaces}

To demonstrate the efficacy of generalized distance metric learning (via manifold-MMC, cf.\ Section \ref{sec:mfd_mmc}) with respect to improving cluster performance, we utilize the \dataset{20newsgroup} dataset, as well as a synthetically generated dataset consisting of points sampled from a helicoid manifold. The results on the synthetic dataset are depicted in Figure \ref{fig:cluster_helicoid}. This dataset was carefully chosen to show the immense potential of considering the an appropriate representation for a given dataset: in the the Euclidean representation, there is no (linear) transformation that can achieve an effective $k$-means clustering, but once the right representation is chosen (the helicoid, in this case), distance metric learning makes the clustering task almost trivial.

This also leads to the question: \emph{how} exactly can clustering be done in a non-Euclidean space?
Note that an arbitrary generalized surface is not necessarily even a vector space (as $S$ is not guaranteed to be closed under vector addition), and thus there does not exist a notion of a \emph{mean vector} or a \emph{center}, a key concept that is required to run the typical $k$-means algorithm. Nevertheless, we can generalize $k$-means in a more natural way. Recall that the $k$-means objective attempts to find a $k$-partition $C_1,\ldots,C_k$ of a given dataset $X = \{x_1,\ldots,x_n\}$
 that minimizes
$$
\sum_{j=1}^k \sum_{i\in C_j} \|x_i - \mu_j \|^2,
$$
where $\mu_j$ is the mean of cluster $C_j$. Since on a generalized surface, we have no concept of the mean, we cannot minimize this form.
To circumvent this, we note that the inside summation can be equivalently rewritten as (for any $j$)
 $$
 \sum_{i\in C_j} \|x_i - \mu_j \|^2 = \frac{1}{2 |C_j|} \sum_{i,i'\in C_j} \|x_i - x_{i'}\|^2.
 $$
This change reformulates the $k$-means optimization solely in terms of pairwise distances \citep{kmeanshardness} and thus extends it to manifolds.  
 
Leveraging this formulation, we present a generalized $k$-means algorithm that can operate on any generalized surface. We define a cluster assignment $\Theta$ to be the assignment of one cluster label from $\{1, ..., k\}$ to each point $x_i$ in our dataset $X = \{x_1, ..., x_n\}$; the value $y_i \in \Theta$ refers to the cluster assignment of the point $x_i$. We define a counting function:
$$K(y_i) = \sum_{j = 1}^n \mathbf{1}[y_i = y_j],$$
which outputs the number of points in $X$ with the same cluster assignment as $x_i$ (including the point $x_i$). The dataset $X$ consists of points embedded in a generalized surface with distance function $\rho$. Thus, the cost of a given cluster assignment $\Theta$ is:
$$C(\Theta; X) = \sum_{i = 1}^n \sum_{j = 1}^n \mathbf{1}[y_i = y_j] \rho(x_i, x_j) \frac{1}{2K(y_i)}.$$
 See Algorithm \ref{alg:cluster} for a detailed implementation. (Notice that the optimization style is akin to Hartigan's method for $k$-means optimization, \citealp{hartigans})

\ificml
\begin{algorithm}[htb!]
\caption{$k$-Means on Generalized Surfaces}
\label{alg:cluster}
\begin{algorithmic}[1]
\STATE $\Theta \leftarrow$ randomly assign each point on the manifold in the dataset $X$ to one of $k$ clusters
\REPEAT
    \FOR{$x_i \in \{x_1, ..., x_n\}$}
    \STATE {current cost = $C(\Theta; X)$}
    \STATE {min cost $\leftarrow$ current cost}
    \FOR{$j \in \{1, ..., k\}$}
    \STATE{Set $\Theta_{new}$ to be equal to $\Theta$ but with the label of $x_i$ set to $j$}
    \STATE{new cost = $C(\Theta_{new}; X)$}
    \IF {new cost $<$ min cost}
    \STATE{$\Theta \leftarrow \Theta_{new}$}
    \ENDIF 
    \ENDFOR
    \ENDFOR
\UNTIL {convergence}
\STATE \textbf{return} $\Theta$
\end{algorithmic}
\end{algorithm}
\fi
For the \dataset{20newsgroup} dataset, we perform the generalized $k$-means clustering on a hyperboloid embedding and a metric learned hyperboloid embedding (via metric-MMC). 
We use normalized mutual information (NMI) to measure the quality of the obtained $20$-way clustering.
Relative to the hyperboloid embedding, the performance of clustering on the metric learned hyperboloid improves by $0.015$ units (or $3\%$ improvement). Given that this dataset has 20 different categories and is thus very difficult to properly cluster, this improvement is significant. 

\ifarxiv
\begin{table*}[tb]
    \centering
        \caption{$k$-nearest neighbor classification results.}
    \label{tab:clf_res}
    \begin{tabular}{|c|c|c|c|c|}
        \hline 
        Dataset & Euclidean & Euclidean+Metric Learn & Hyperbolic & Hyperbolic+Metric Learn \\
        \hline \hline
 \dataset{football} & 0.41 $\pm$	0.01 & 0.40 $\pm$ 0.09 & 0.29 $\pm$ 0.09 & \textbf{0.25 $\pm$ 0.10} \\
  \dataset{polbooks} & 0.24 $\pm$	0.05 & 0.31 $\pm$ 0.12 & 0.25 $\pm$ 0.06 & \textbf{0.23 $\pm$ 0.06} \\
  \dataset{adjnoun} & 0.58 $\pm$	0.06 & 0.56 $\pm$ 0.07 & 0.55 $\pm$ 0.09 & \textbf{0.49 $\pm$ 0.05} \\
\hline
    \end{tabular}
\end{table*}
\fi

\subsection{Classification on the Hyperboloid}

To demonstrate the efficacy of generalized distance metric learning (via manifold-LMNN, cf.\ Section \ref{sec:mfd_lmnn}) with respect to improving classification performance, we use the \dataset{football}, \dataset{polbooks}, and \dataset{adjnoun} datasets. Note that the generalized surface we have chosen for each of these datasets is the two-dimensional hyperboloid. 

For each dataset, the classifier performance is measured using the standard 0-1 error, and the results are presented in Table \ref{tab:clf_res}. In all cases, the classification performance on the metric-learned hyperboloid is significantly better than in the other three embedding options. This emphasizes that the right notion of distance and an appropriate (perhaps non-Euclidean) choice of representation is sometimes crucial to attain good performance.

In particular, we note that linear classifier performance on the \dataset{football} dataset (using a one-vs-rest strategy) is particularly poor, with best reported error (as per \citealp{hgeom_svm}) being $0.79$ on a Euclidean embedding, with an slight improvement to $0.76$ on a hyperboloid embedding. In contrast, our $k$-nearest neighbor error on the metric-learned hyperboloid is $0.25 \pm 0.10$, a significant improvement. 






\nocite{manifold_ml}

\bibliography{hml_icml19}
\bibliographystyle{icml2019}

\clearpage

\onecolumn
\appendix

\section{Derivation of Arc Length Integrand for Hyperboloid}
\label{app:derive_hyperbolic_path_approx}

Take the two-dimensional hyperboloid $H$, which has base space $B = \mathbb{R}^2$ and diffeomorphism $F(x) = (x,\sqrt{1 + x^{\mathsf{T}} x})$ for $x \in B$. Let $\kappa(t)$ be any path (parameterized from $t = 0$ to $1$) from $a$ to $b$ in $B$. Recall from Section \ref{sec:dist_over_S} that the length of the path $\kappa$ mapped to the hyperboloid is:

$$\LEN_F[\kappa] = \int_0^1 \sqrt{ \Big\langle \frac{d}{dt}F(\kappa(t)), \frac{d}{dt}F(\kappa(t)) \Big\rangle} \; dt.$$

We want to compute $\frac{d}{dt} F(\kappa(t))$. By the chain rule of differentiation, this is equivalent to $\frac{dF}{d\kappa} \frac{d\kappa}{dt}$. The derivative $\frac{dF}{d\kappa}$ can be represented as a $3 \times 2$ matrix, which we will denote as $D(t)$:

$$D(t) := \begin{bmatrix} I_{2\times 2}  \vspace{0.1in} \\ K(t) \end{bmatrix} =  \begin{bmatrix} 1 & 0 \\ 0 & 1 \\ K(t)_1 & K(t)_2  \end{bmatrix},$$

where $K(t) = \frac{1}{\sqrt{1 + \kappa(t)^{\mathsf{T}} \kappa(t)}} \kappa(t)$ is a vector tangent to the base space $B$ (at $t$). For notational convenience, let $\dot{\kappa}(t)$ denote the derivative $\frac{d}{dt} \kappa(t)$. Then, we have:

\begin{align*}
\sqrt{ \Big\langle \frac{d}{dt}F(\kappa(t)), \frac{d}{dt}F(\kappa(t)) \Big\rangle} &= \sqrt{\dot{\kappa}^{\mathsf{T}} D^\mathsf{T} G D \dot{\kappa}}
= \sqrt{\dot{\kappa}^{\mathsf{T}} \dot{\kappa} - \frac{(\kappa^{\mathsf{T}} \dot{\kappa})^2}{1 + \kappa^{\mathsf{T}} \kappa}},
\end{align*}
where the first equality is by noting that the inner product is with respect to the Minkowski metric tensor endowed on the hyperboloid, and is encoded by $G := \Big[\begin{array}{cc} I_{2\times 2} & 0\\ 0 & -1 \end{array} \Big] $ (cf.\ Section \ref{sec:example_instances} and \citealp{hgeom_reynolds} for more details).

Therefore, for the hyperboloid, 

$$\LEN_F[\kappa] = \int_0^1 \sqrt{\dot{\kappa}^{\mathsf{T}} \dot{\kappa} - \frac{(\kappa^{\mathsf{T}} \dot{\kappa})^2}{1 + \kappa^{\mathsf{T}} \kappa}} \;\;dt.$$

Let us denote the above integrand by $V := \sqrt{\dot{\kappa}^{\mathsf{T}} \dot{\kappa} - \frac{(\kappa^{\mathsf{T}} \dot{\kappa})^2}{1 + \kappa^{\mathsf{T}} \kappa}}$. Note that $V$ is only a function of $\kappa$ and $\dot{\kappa}$, and therefore, the solution to the Euler-Lagrange equation:
$$\frac{\partial V}{\partial \kappa} - \frac{d}{dt} \frac{\partial V}{\partial \dot{\kappa}} = 0$$
yields the optimal path $\kappa^\ast$ that minimizes $\LEN$, and therefore yields the exact distance function on $H$.

\clearpage

\section{Proof of Theorem \ref{lm:unif_conv_all}}
\label{app:proofs}
%
%

Following a setup similar to \citet{mlrn_verma_samplecomplexity}, let $\mathcal{P}$ be the probability measure induced by the random variable
$(\mathbf{X}, Y)$, where $\mathbf{X}:= (x,x')$, $Y:=\indicate[y=y']$, st. $((x,y),(x',y')) \sim (\D \times \D)$.

Define function class 
\begin{align*}
& \mathcal{H} := 
 \Bigg\{ h_L^F \!: \mathbf{X} \mapsto \rho_L^F(x,x') \Bigg|\!  \begin{array}{c} L \in \mathsterling \\ \mathbf{X} = (x,x') \in (S \times S) \end{array} \!\! \Bigg\},
\end{align*}
and consider any loss function $\phi^\lambda(\rho, Y)$ that is $\lambda$-Lipschitz in the first argument.
Then, we are interested in bounding the quantity 
\begin{align*}
\sup_{h_L^F \in \mathcal{H}} \E_{(\mathbf{X},Y)\sim \mathcal{P}} [\phi^\lambda(h_L^F(\mathbf{X}), Y)] - \frac{1}{m} \sum_{i=1}^m \phi^\lambda(h_L^F(\mathbf{X}_i),Y_i),
\end{align*}
where $\mathbf{X}_i:=(x_{1,i},x_{2,i}) $, $Y_i:=\indicate[y_{1,i}=y_{2,i}]$
from the paired sample $S_m^{\textup{pair}} = \{((x_{1,i},y_{1,i}),(x_{2,i},y_{2,i}))
\}_{i=1}^m$ derived from the given sample $Z_m$. 

Define $b_{j,i}$ be such that $F(b_{j,i}) = x_{j,i}$ for $1\leq i\leq m$ and $j\in\{1,2\}$, and define $\bar{b}_i := b_{1,i} - b_{2,i}$ for each $ \mathbf{X}_i=(x_{1,i},x_{2,i})$. 
Then, the Rademacher complexity\footnote{See the definition of Rademacher complexity in the statement of Lemma \ref{lm:rad_complexities_unif_bound}.} of our function class $\mathcal{H}$ (with respect to the distribution $\mathcal{P}$) is bounded, since (let $\sigma_1,\ldots, \sigma_m$ denote
independent uniform $\{\pm 1\}$-valued random variables)
\begin{align*}
\mathcal{R}_m(\mathcal{H}, \mathcal{P}) 
& := \E_{\substack{\mathbf{X}_i,\sigma_i \\ i\in[m]}}  \Bigg[ \sup_{h_L^F \in \mathcal{H}} \frac{1}{m} \sum_{i=1}^m \sigma_i h_L^F(\mathbf{X}_i) \Bigg]  \\
&\leq \frac{C_F}{m} \cdot \E_{\substack{\mathbf{X}_i,\sigma_i \\ i\in[m]}} \sup_{L \in \mathsterling} \Big[ \sum_{i=1}^m \sigma_i \bar{b}_i^\mathsf{T} L^\mathsf{T}L \bar{b}_i   \Big] 
\\
&
= \frac{C_F}{m} \cdot \E_{\substack{\mathbf{X}_i,\sigma_i \\ i\in[m]}} \sup_{\substack{L \in \mathsterling, \textrm{ s.t.} \\ [a^{jk}]_{jk} := L^\mathsf{T}L }} \Bigg[ \sum_{j,k} a^{jk} \sum_{i=1}^m \sigma_i \bar{b}_i^j \bar{b}_i^k \Bigg] \\
&\leq \frac{C_F }{m} \cdot \E_{\substack{\mathbf{X}_i,\sigma_i \\ i\in[m]}} \sup_{L \in \mathsterling} \Bigg[ \|L^\mathsf{T}L\|_{_\textrm{fro}} \Bigg( \sum_{j,k} \Big(\sum_{i=1}^m \sigma_i \bar{b}_i^j \bar{b}_i^k \Big)^2 \Bigg)^{1/2} \Bigg] \\
&\leq \frac{C_F \cdot C_L}{m} \cdot \E_{\substack{\mathbf{X}_i,  i\in[m]}} \Bigg( \E_{\substack{\sigma_i,i\in[m]}}  \sum_{j,k} \Big(\sum_{i=1}^m \sigma_i \bar{b}_i^j \bar{b}_i^k \Big)^2 \Bigg)^{1/2} \\
&= \frac{C_F \cdot C_L}{m} \cdot \E_{\substack{\mathbf{X}_i,  i\in[m]}} \Bigg( \sum_{j,k} \sum_{i=1}^m \Big(\bar{b}_i^j \Big)^2 \Big(\bar{b}_i^k \Big)^2 \Bigg)^{1/2} \\
&= \frac{C_F \cdot C_L}{m} \cdot \E_{\substack{\mathbf{X}_i,  i\in[m]}} \Bigg( \sum_{i=1}^m \|\bar{b}_i\|^4 \Bigg)^{1/2} \\
&= \frac{C_F \cdot C_L}{m} \cdot \E_{\substack{ (x_i, x'_i) \sim (\D|_S \times \D|_S) ,\\  i\in[m]}} \Bigg( \sum_{i=1}^m \|b_i - b'_i\|^4 \Bigg)^{1/2} \\
&\leq \frac{C_F \cdot C_L}{\sqrt{m}} \cdot \Bigg (\E_{\substack{ (x, x') \sim (\D|_S \times \D|_S)}} \|b - b'\|^4 \Bigg)^{1/2} \\
&\leq 4 {C_F \cdot C_L \cdot C_B} / { \sqrt{m}}. 
\end{align*}

Recall that $\D$ has bounded support induced on $B$ (with bound $C_B$). Thus, by noting that $\phi^\lambda$ is $O(C_L C_F C_B)$ bounded function that is $\lambda$-Lipschitz in the first argument, we can apply Lemma \ref{lm:rad_complexities_unif_bound} and get the desired 
uniform deviation bound. \qed

\begin{lemma} \textbf{\emph{[Rademacher complexity of bounded Lipschitz loss functions \cite{lt:bartlett_mendelson_radgauss_complexities}] } }
\label{lm:rad_complexities_unif_bound}
Let $\D$ be a fixed unknown distribution over $X \times \{-1,1\}$, and
let $S_m$ be an i.i.d.\ sample of size $m$ from $\D$.
Given
a hypothesis class $\mathcal{H}\subset\R^{X}$ and a loss function 
$\ell : \R \times \{-1,1\} \rightarrow \R$, such that 
$\ell$ is $c$-bounded, and is $\lambda$-Lipschitz in the first argument, that
is, $\sup_{(y',y)\in\R\times\{-1,1\}} | \ell(y',y) | \leq c$, and $|\ell(y',y)
- \ell(y'',y)|\leq \lambda |y' - y''|$, we have the following:

for any $0<\delta<1$, with probability at least $1-\delta$, every $h\in \mathcal{H}$ satisfies
\begin{equation*}
\err(\ell \circ h,\D) \leq \err(\ell \circ h,S_m) + 2\lambda \mathcal{R}_m(\mathcal{H}, \D) + c\sqrt{\frac{2 \ln(1/\delta)}{m}},
\end{equation*}
where 
\begin{itemize}
\item $\err(\ell \circ h,\D) := \E_{{x,y}\sim \D} [\ell(h(x),y)]$,
\item $\err(h,S_m) := \frac{1}{m} \sum_{(x_i,y_i)\in S_m} \ell(h(x_i),y_i)   $,
\item 
$\mathcal{R}_m(\mathcal{H}, \D)$ is the Rademacher complexity of the function class $\mathcal{H}$ with respect to the distribution
$\D$ given $m$ i.i.d.\ samples, and is defined as:
\begin{equation*}
\mathcal{R}_m(\mathcal{H},\D) := \E_{\substack{ x_i\sim \D|_X, \\ \sigma_i \sim \mathrm{unif}\{\pm 1\}, \\ i\in[m] }} \Bigg[\sup_{{h}\in\mathcal{H}} \frac{1}{m} \sum_{i=1}^m \sigma_i h(x_i)   \Bigg],
\end{equation*}
where $\sigma_i$ are independent uniform $\{\pm 1\}$-valued random variables.
\end{itemize}
\end{lemma}